\documentclass{article}

\usepackage{arxiv}

\usepackage[utf8]{inputenc} 
\usepackage[T1]{fontenc}    
\usepackage{hyperref}       
\usepackage{url}            
\usepackage{booktabs}       
\usepackage{amsfonts}       
\usepackage{nicefrac}       
\usepackage{microtype}      
\usepackage{cleveref}       
\usepackage{lipsum}         
\usepackage{graphicx}
\usepackage{doi}

\RequirePackage{tikz}
\usepackage[authoryear]{natbib}
\usepackage{subcaption}
\usepackage{pgfplots}
\usepackage{tikz}
\usetikzlibrary {arrows.meta,automata,positioning,shapes.geometric}
\usepackage{algorithm, algpseudocode}

\usepackage{multirow}
\usepackage{float}
\usepackage[utf8]{inputenc}
\title{A knowledge-driven AutoML architecture}

\author{Corneliu, Cofaru\\
	Vrije Universiteit Brussel\\
	Pleinlaan 9, 1050 Brussels, Belgium \\
	\texttt{corneliu.cofaru@vub.be} \\
	\And
	Johan, Loeckx\\
	Vrije Universiteit Brussel,\\
	Pleinlaan 9, 1050, Brussels, Belgium\\
	\texttt{johan.loeckx@vub.be} \\
}

\renewcommand{\shorttitle}{A knowledge-driven AutoML architecture}

\hypersetup{
pdftitle={A knowledge-driven AutoML architecture},
pdfsubject={cs.AI, cs.LG, cs.SE},
pdfauthor={Corneliu Cofaru, Johan Loeckx},
pdfkeywords={Knowledge representation, knowledge engineering, AutoML, machine learning},
}

\begin{document}
\maketitle

\begin{abstract}
    This paper proposes a knowledge-driven AutoML architecture for pipeline and deep feature synthesis. The main goal is to render the AutoML process explainable and to leverage domain knowledge in the synthesis of pipelines and features. The architecture explores several novel ideas: first, the construction of pipelines and deep features is approached in an unified way. Next, synthesis is driven by a shared knowledge system, interactively queried as to what pipeline operations to use or features to compute. Lastly, the synthesis processes takes decisions at runtime using partial solutions and results of their application on data. Two experiments are conducted to demonstrate the functionality of a na\"{\i}ve implementation of the proposed architecture and to discuss its advantages, trade-offs as well as future potential for AutoML.
\end{abstract}


\section{Introduction}\label{1_introduction}

Automated machine learning (AutoML) gathered a significant amount of attention in recent years as a way of automating some of the typical workflows in machine learning ({ML}) and data science more broadly. For a comprehensive and systematic view on the subject, there is an already growing number of survey works that cover the state-of-the-art~\cite{Hutter2019, Yao2018, Elshawi2019, Zoller2021, Truong2019, He2021, Hospedales2020, Vanschoren2018, Karmaker2021}. Currently, it is becoming apparent that the size of the potential problem space, required solution sophistication, transparency and legal constraints~\cite{Roscher2020, Drozdal2020, Rudin2021, Veale2021, Smuha2021} render AutoML a problem extremely difficult to define and solve either holistically or agnostically. The first translates into finding an optimal compromise between multiple criteria such as accuracy, computational resources or transparency while the latter into handling correctly and in a general fashion data semantics and problem-specific constraints.

Our motivation is set in the context of an R\&D project in which we focused on the investigation of transparency, explain-ability and audit-ability in AutoML and on rendering domain knowledge available and (re)usable by data scientists. These are known key challenges in AutoML state-of-the-art~\cite{Karmaker2021, Elshawi2019} and, to the extent of our knowledge, have received little attention in extant work. The end-result of following the motivation lines above is an AutoML architecture\footnote{The library is available at \texttt{https://github.com/zgornel/Kdautoml.jl} and the experiments in this paper at \texttt{https://github.com/zgornel/knowledge-driven-automl}.}, characterized by three design principles:
\begin{itemize}
    \item \textbf{Data-driven}: computer code snippets, associated symbols and links between them are stored in and accessed through a knowledge system. This enables domain knowledge representation as well as explain-ability, through the analysis of communication and reasoning of the knowledge system.
    \item \textbf{Unified architecture}: pipeline and deep feature synthesis use a common knowledge system and share the architectures' functional organization.
    \item \textbf{Runtime dynamic}: the architecture allows runtime access to partial pipelines or features and their data outputs, feature values and models, allowing for a wider range of domain knowledge to be expressed.
\end{itemize}

Deep feature synthesis~\cite{Kanter2015} represents a suitable choice for feature engineering as its 'deep' or recursive nature and feature types allows for both the calculation of a wide variety of data processing operations and straightforward integration of domain knowledge. At its core it is a problem very similar to pipeline synthesis, in that features can be built out of combinations of more basic components given certain implicit or explicit rules and conditions under which this can occur.

The architectural design principles presented here focus on 'traditional' AutoML and not neural architecture search (NAS)~\cite{He2021, Peng2020}. At this point it is difficult to state what such an application may entail.

The rest of this paper is organized as follows: Section~\ref{sec:Architecture} presents the proposed architecture, Section~\ref{sec:Implementation} describes our reference implementation which is employed in Section~\ref{sec:ExperimentalResults} to perform an early experimental validation. Lastly, Sections~\ref{sec:Discussion} and~\ref{sec:Conclusion} contain the discussion and conclusions.

\section{Architecture}\label{sec:Architecture}
Some of the core concepts of this paper can be individually identified in recent works. ML pipeline synthesis knowledge was represented as grammars \cite{DeSa2017, Assuncao2020}, hierarchical task networks \cite{Mohr2018}, rules \cite{Shang2019}, optimization constraints \cite{Liu2019} and ML~models \cite{Yakovlev2020, Yang2020}. Complex pipeline structures are the focus of \cite{Nikitin2021} while in \cite{Yakovlev2020} pipeline synthesis uses partial pipeline outputs. Explain-ability and use of logical rules in the pipeline generation process has been recently explored in~\cite{Das2020}. Deep or recursive feature engineering for relational datasets is developed in~\cite{Kanter2015, Lam2017} and knowledge integration in feature synthesis is the focus of~\cite{Galhotra2019}.

Our work attempts to generalize to a certain extent over all of these works through an explicit separation and representation of an AutoML's system functional components. We draw inspiration from previous research in blackboard architectures \cite{Carver1994} and procedural semantics \cite{Johnson-Laird1977} to create, what could be viewed as, a code-generating complement to knowledge-driven ML code mining and analysis \cite{Patterson2017, Patterson2018}.
\begin{figure}[ht]
    \centering
        \begin{tikzpicture}[scale=0.8, every node/.style={scale=0.8}]
            \node [draw,
                inner sep=5pt,
                thick,
                minimum width=2cm,
                minimum height=1.5cm,
                align=center
            ]  (cf) at (0,0) {Control Flow\\(CF)};

            \node [draw,
                inner sep=5pt,
                thick,
                minimum width=2cm,
                minimum height=1.5cm,
                left=2.5cm of cf,
                align=center,
            ] (pe) {Program\\Execution (PE)};

            \node [draw,
                inner sep=5pt,
                thick,
                minimum width=2cm,
                minimum height=1.5cm,
                right=2.5cm of cf,
                align=center,
            ] (ks) {Knowledge\\System (KS)};

            \path[->,arrows = {-Latex[open]}] (cf.10)  edge      node[above, sloped]  {(1) Query}     (ks.170)
                      (ks.190)  edge      node[below, sloped]  {(2) Reply}     (cf.350)
                      (cf.170)  edge      node[above, sloped]  {(3) Execute}   (pe.9)
                      (pe.351)  edge      node[below, sloped]  {(4) Results}   (cf.190);

                \draw[->,arrows = {-Latex[open]},dashed] (cf.45) to[out=90,in=90] node [sloped,above] {Synthesis data} (ks.90);
                \draw[->,arrows = {-Latex[open]},dashed] (ks.south) to[out=270,in=270] node [sloped,below] {Operation nodes} (cf.315);
                \draw[arrows = {Latex[open]-Latex[open]},dashed] (cf.225) to[out=270,in=270] node (istext) [sloped,below] {Intermediary solution} (pe.south);
             \node [above=1.5cm of cf] (cfnorth) {Input};
             \draw[->,arrows = {-Latex[open]},dashed] (cfnorth) -- (cf.north);
             \node [below=1.5cm of cf] (cfsouth) {Output};
             \draw[->,arrows = {-Latex[open]},dashed] (istext.east) to[out=0, in=90] (cfsouth.north);

		\end{tikzpicture}
    \caption{The knowledge-driven AutoML architecture comprises three functionally distinct blocks which model the synthesis processes (CF), represent and render accessible pipeline and feature component information (KS) and assemble and execute generated code (PE). Solid arrows indicate operation flow, dashed arrows data flow and numbers the order in the synthesis process respectively.}\label{img:Architecture}
\end{figure}
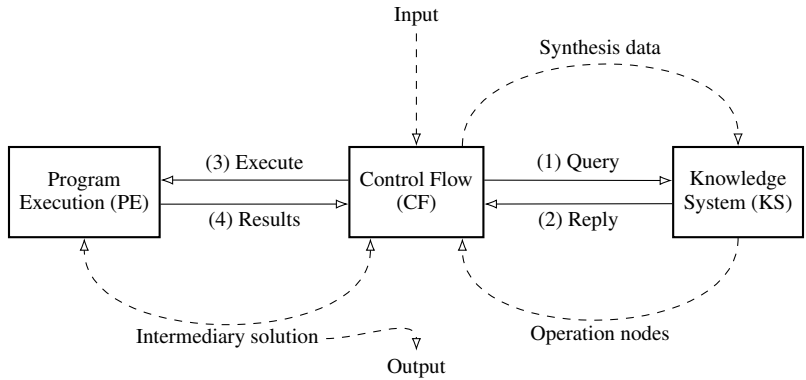

In the following subsections we shall discuss the three main functional components of the architecture which are shown in Figure~\ref{img:Architecture}.

\subsection{Knowledge system (KS)}\label{ssec:KnowledgeSystem}

\subsubsection{Knowledge representation}
At the highest level of conceptualization, we assume pipelines represented by directed acyclic graphs (DAG) and features by abstract syntax trees (AST). We shall refer to the basic building block of both structures as \textit{node}. It consists of a symbolic name that uniquely identifies it and computer code that implements its functionality. Nodes can be connected to each other, process and transmit data. To represent pipeline and feature knowledge, we chose knowledge graphs as this allows for a straightforward representation of node information and, through different types of links, shared node commonalities and conditions related to their practical use. This allows to abstractly represent in a unified manner ML pipeline operations, data features, feature components and conditions over them.
\begin{figure}[t]
    \begin{subfigure}{\textwidth}
        \centering
        \begin{tikzpicture}
            [
            level 1/.style = {line width = 0.5pt, level distance=1.1cm, sibling distance = .5cm},
            level 2/.style = {line width = 0.5pt, level distance=1.1cm, sibling distance = 1.5cm},
            level 3/.style = {line width = 0.5pt, level distance=1.1cm, sibling distance = 1cm},
            edge from parent path =  {(\tikzparentnode\tikzparentanchor) .. controls +(0,-0.5) and +(0,0.5) .. (\tikzchildnode\tikzchildanchor)}]
            \node (nodea) [line width=0.7pt, circle, draw, double, minimum size = 6pt] {}  
                child {node [circle,line width=0.7pt, draw, minimum size = 6pt] {} edge from parent[to-]
                    child {node (nodec) [circle,line width=0.7pt,draw,fill, minimum size = 6pt] {}
                        child {node (nodep1) [draw,minimum size = 6pt] {} edge from parent [-to,dashed]}
                        child {node [minimum size = 6pt] {...} edge from parent [draw=white]}
                        child {node (nodep2) [draw,minimum size = 6pt] {} edge from parent [-to,dashed]}
                        }
                    child {node [line width=0.7pt] {...} edge from parent [draw=white]}  
                    child {node [circle,line width=0.7pt,draw,fill,minimum size = 6pt] {} edge from parent[to-]
                        child {node [draw,minimum size = 6pt] {} edge from parent [-to,dashed]}
                        child {node [draw,minimum size = 6pt] {} edge from parent [-to,dashed]}
                        child {node [minimum size = 6pt] {...} edge from parent [draw=white]}
                        child {node (nodep3) [draw,minimum size = 6pt] {} edge from parent [-to,dashed]}
                        child {node (nodep4) [draw,minimum size = 6pt] {} edge from parent [-to,dashed] node[minimum size=6pt] (npcondby) {}}
                        }
                    edge from parent[to-] node[minimum size=6pt] (nisa) {}
                }
                child {node [line width=0.7pt,]{...} edge from parent [draw=white] }  
                child {node [circle,line width=0.7pt,draw,minimum size = 6pt] {} edge from parent[to-]}
                child {node [line width=0.7pt,]{...} edge from parent [draw=white] }
                child {node [circle,line width=0.7pt,draw,minimum size = 6pt] {} edge from parent[to-]}
                child {node [line width=0.7pt,]{...} edge from parent [draw=white] }
                child {node [circle,line width=0.7pt,draw,minimum size = 6pt] (nodecomp) {} edge from parent[to-]}; 

            \node [right=9pt of nodea] (antext) {\small Abstract node (top)};
            \draw [-Latex,thin] (antext.west) -- (nodea.east);

            \node [right=9pt of nodecomp,align=center] (compntext) {\small Abstract nodes\\\small(pipeline components)};
            \draw [-Latex,thin] (compntext.west) -- (nodecomp.east);

            \node [left=9pt of nodec] (cntext) {\small Operation nodes};
            \draw [-Latex,thin] (cntext.east) -- (nodec.west);

            \node [left=9pt of nisa] (nisatext) {\small{Hierarchical links}};
            \draw [-Latex,thin] (nisatext.east) -- (nisa.west);

            \node [right=9pt of npcondby] (npcondbytext) {\small{Precondition links}};
            \draw [-Latex,thin] (npcondbytext.west) -- (npcondby.east);

            \node [left=9pt of nodep1] (nodep1text) {\small{Preconditions}};
            \draw [-Latex,thin] (nodep1text.east) -- (nodep1.west);

        \end{tikzpicture}
        \caption{For pipeline synthesis, abstract nodes are organized hierarchically with operation nodes as terminals of the hierarchy, linked to preconditions.}
        \label{img:ps_knoweledge_representation}
    \end{subfigure}
    \hfill
    \begin{subfigure}{\textwidth}
        \centering
        \begin{tikzpicture}
            [
            level 1/.style = {line width = 0.5pt, level distance=1.1cm, sibling distance = .5cm},
            level 2/.style = {line width = 0.5pt, level distance=1.1cm, sibling distance = .5cm},
            level 3/.style = {line width = 0.5pt, level distance=1.1cm, sibling distance = 1.5cm},
            level 4/.style = {line width = 0.5pt, level distance=1.1cm, sibling distance = 1cm},
            edge from parent path =  {(\tikzparentnode\tikzparentanchor) .. controls +(0,-0.5) and +(0,0.5) .. (\tikzchildnode\tikzchildanchor)}]
            \node (nodea) [line width=0.7pt, circle, double, draw, minimum size = 6pt] {}  
                child {node [circle,line width=2pt, draw, minimum size = 6pt] (nodef1){}
                    child {node [circle,line width=0.7pt, draw, minimum size = 6pt] {} edge from parent[to-,line width=.5pt,double] node[minimum size=6pt] (nhasa) {}}
                    child {node [line width=0.7pt,]{...} edge from parent [draw=white] }
                    child {node [circle,line width=0.7pt, draw, minimum size = 6pt] (nodefc1) {} edge from parent[to-,line width=.5pt,double] 
                        child {node (nodec) [circle,line width=0.7pt,draw,fill, minimum size = 6pt] {}
                            child {node (nodep1) [draw,minimum size = 6pt] {} edge from parent [-to,dashed]}
                            child {node [minimum size = 6pt] {...} edge from parent [draw=white]}
                            child {node (nodep2) [draw,minimum size = 6pt] {} edge from parent [-to,dashed]}
                            }
                        child {node [line width=0.7pt] {...} edge from parent [draw=white]}  
                        child {node [circle,line width=0.7pt,draw,fill,minimum size = 6pt] {}
                            child {node [draw,minimum size = 6pt] {} edge from parent [-to,dashed]}
                            child {node [draw,minimum size = 6pt] {} edge from parent [-to,dashed]}
                            child {node [minimum size = 6pt] {...} edge from parent [draw=white]}
                            child {node (nodep3) [draw,minimum size = 6pt] {} edge from parent [-to,dashed]}
                            child {node (nodep4) [draw,minimum size = 6pt] {} edge from parent [-to,dashed] node[minimum size=6pt] (npcondby) {}}
                            }
                        edge from parent[to-]
                    }
                   edge from parent[to-] node (nisa){}
                }
                child {node [line width=0.7pt,]{...} edge from parent [draw=white] }
                child {node [circle,line width=2pt,draw,minimum size = 6pt] (nodef2) {} edge from parent[to-]}
                child {node [line width=0.7pt,]{...} edge from parent [draw=white] }
                child {node [circle,line width=2pt,draw,minimum size = 6pt] (nodef3) {} edge from parent[to-]}
                child {node [line width=0.7pt,]{...} edge from parent [draw=white] }
                child {node [circle,line width=2pt,draw,minimum size = 6pt] (nodef4){} edge from parent[to-]
                    child {node [circle,line width=0.7pt,draw,minimum size = 6pt] (nodefc41) {} edge from parent[to-,line width=.5pt,double]}
                    child {node [line width=0.7pt,minimum size = 6pt] {...} edge from parent [draw=white]}
                    child {node [circle,line width=0.7pt,draw,minimum size=6pt] (nodefc42) {} edge from parent[to-,line width=.5pt,double]}
                    edge from parent [to-]
                };

            \draw [to-, line width=0.5pt,double] (nodef2.south)  .. controls +(0,-0.5) and +(0,0.5) ..  (nodefc1.north);
            \draw [to-, line width=0.5pt,double] (nodef3.south)  .. controls +(0,-0.5) and +(0,0.5) ..  (nodefc1.north);
            \draw [to-, line width=0.5pt,double] (nodef3.south)  .. controls +(0,-0.5) and +(0,0.5) ..  (nodefc42.north);

            \node [right=9pt of nodea] (antext) {\small Abstract node (top)};
            \draw [-Latex,thin] (antext.west) -- (nodea.east);

            \node [right=9pt of nodef4, align=center] (f4ntext) {\small Abstract nodes\\\small(feature types)};
            \draw [-Latex,thin] (f4ntext.west) -- (nodef4.east);

            \node [right=9pt of nodefc42, align=center] (fc42ntext) {\small Abstract nodes\\\small(feature components) };
            \draw [-Latex,thin] (fc42ntext.west) -- (nodefc42.east);

            \node [left=9pt of nodec] (cntext) {\small Operation nodes};
            \draw [-Latex,thin] (cntext.east) -- (nodec.west);

            \node [left=9pt of nisa] (nisatext) {\small{Hierarchical links}};
            \draw [-Latex,thin] (nisatext.east) -- (nisa.west);

            \node [left=9pt of nhasa] (nhasatext) {\small{Feature AST links}};
            \draw [-Latex,thin] (nhasatext.east) -- (nhasa.west);

            \node [right=9pt of npcondby] (npcondbytext) {\small{Precondition links}};
            \draw [-Latex,thin] (npcondbytext.west) -- (npcondby.east);

            \node [left=9pt of nodep1] (nodep1text) {\small{Preconditions}};
            \draw [-Latex,thin] (nodep1text.east) -- (nodep1.west);
        \end{tikzpicture}
        \caption{For feature synthesis, the abstract nodes representing features are linked to the ones representing feature components. The feature AST links that connect the two indicate which components are part of a specific feature AST.}
        \label{img:fs_knoweledge_representation}
    \end{subfigure}
    \caption{The pipeline, feature and feature AST component nodes are represented in a knowledge graph in which vertices correspond to nodes and edges to links. The three different link types form node hierarchies, indicate what feature components belong to each feature type and associate preconditions to nodes respectively.}\label{img:knowledge_representation}
\end{figure}
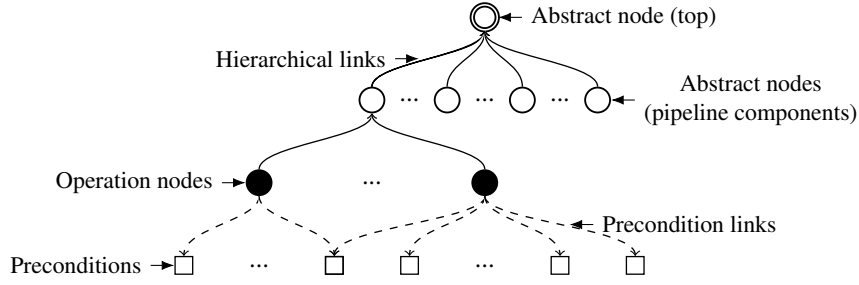
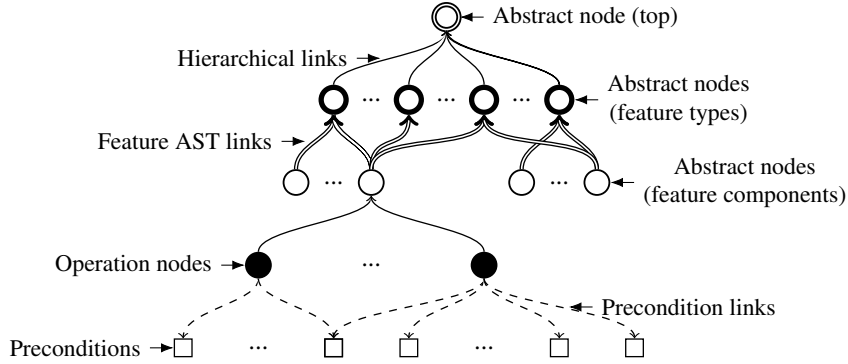
The knowledge graph structure employed here, schematically shown in Figure~\ref{img:knowledge_representation}, is made up of three node types and three link types:
\begin{itemize}
    \item \textbf{Abstract nodes} - represent higher-level functional abstractions of pipeline operations, features and features components;
    \item \textbf{Operation nodes} - represent computational operations that are performed in pipelines and features. For pipelines, these correspond to ML algorithms and data processing functions while in features to data filtering, grouping and processing respectively;
    \item \textbf{Preconditions} - are special nodes whose code is executed during synthesis;
    \item \textbf{Hierarchical links} - allow for a hierarchical organization of nodes, with operation nodes at the bottom and every node linked to at least one other higher abstraction node;
    \item \textbf{Feature AST links} (feature synthesis only) - indicate which feature components belong to which feature, with every component linked to at least one feature type;
    \item \textbf{Precondition links} - link each precondition to at least one operation.
\end{itemize}
The knowledge structure gradually evolved into this form during the development of the system. For pipeline synthesis, the node hierarchy allows control over the number of candidates that can be added to the solution during synthesis and, through this, the dimension of the solution search space. In feature synthesis, the feature AST links associate (hierarchies of) feature components to feature types. The preconditions serve in both cases to reduce the search space by enabling or disabling specific nodes for synthesis use.

\subsubsection{Knowledge querying}
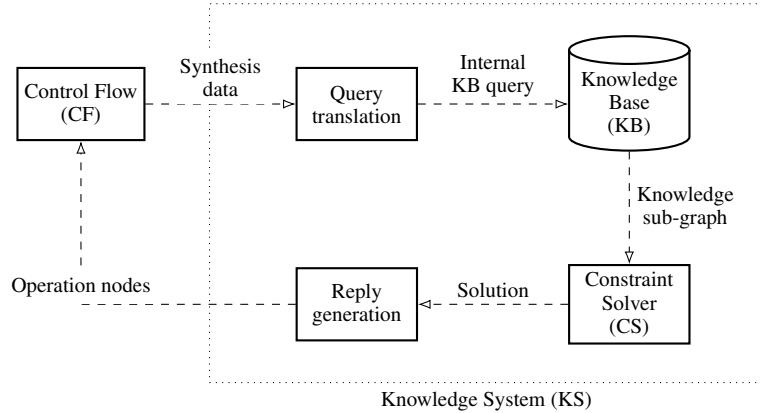
\begin{figure}[ht]
    \centering
        \begin{tikzpicture}[scale=0.8, every node/.style={scale=0.8}]
            \node [draw,
                thick,
                minimum width=2cm,
                minimum height=1.2cm,
                align=center
            ]  (controlflow) at (0,0) {Control Flow\\(CF)};

            \node [draw,
                thick,
                minimum width=2cm,
                minimum height=1.2cm,
                right=2cm of controlflow,
                align=center
            ] (querytranslation) {Query\\translation};

            \node [draw,
                cylinder,
                aspect=0.25,
                shape border rotate=90,
                thick,
                minimum width=2cm,
                minimum height=1.2cm,
                right=2cm of querytranslation,
                align=center
            ] (kb) {Knowledge\\Base\\(KB)};

            \node [draw,
                thick,
                minimum width=2cm,
                minimum height=1.2cm,
                below=1.5cm of kb,
                align=center
            ] (cs) {Constraint\\Solver\\(CS)};

            \node [draw,
                thick,
                minimum width=2cm,
                minimum height=1.2cm,
                left=2cm of cs,
                align=center
            ] (replygeneration) {Reply\\generation};

            \node [draw,
                dotted,
                minimum width=9.2cm,
                minimum height=6.3cm,
                above right=1.2 of controlflow,
                anchor=north west,
            ] (ksd) {};

            \draw[-{Latex[open]},dashed] (controlflow.east) -- (querytranslation.west)
                node[midway,above,align=center,fill=white] (arrow) {Synthesis\\data};

            \node [
                minimum width=0.1cm,
                minimum height=0.1cm,
                below=0.01of ksd,
                align=center,
            ] (kst) {Knowledge System (KS)};
            \draw[-{Latex[open]},dashed] (querytranslation.east) -- (kb.west)
                node[midway,above,align=center]{Internal\\KB query};

            \draw[-{Latex[open]},dashed] (kb.south) -- (cs.north)
                node[midway,right,align=center]{Knowledge\\sub-graph};

            \draw[-{Latex[open]},dashed] (cs.west) -- (replygeneration.east)
                node[midway,above,align=center]{Solution};

            \draw[-{Latex[open]},dashed] (replygeneration.west) -| (controlflow.south)
                node[midway,above,align=center,fill=white]{Operation nodes};
		\end{tikzpicture}
    \caption{The internal Knowledge System (KS) components and data flows involved in answering Control Flow (CF) queries.}\label{img:knowledge_system}
\end{figure}
The role of the knowledge system, schematically shown in more detail in Figure~\ref{img:knowledge_system}, is to inform control flow what pipeline operation to add to the current pipeline or features to the feature set respectively. Two main components can be distinguished: a \textit{knowledge base} holding knowledge graph data and a \textit{constraint solver} that executes precondition code and performs operation node selection. Queries contain data that reflects the state of the synthesis process while replies contain pipeline operation nodes or lists of operation nodes for feature construction. To determine if an operation node can be used, a conjunction of all linked precondition code return values is calculated.

Query answering, sketched in Function~\ref{alg:ks_query}, starts with the reception of synthesis data which is transformed into a query for the knowledge base. The reply from the knowledge base, a sub-graph of the full data, is passed to the constraint solver. The solver builds a constraint program in a manner described by Function~\ref{alg:cs_build_solve} and solves it. Before passing the response back to control flow, node data is added back to the constraint solver solution symbols.

\subsection{Control flow (CF)}\label{ssec:ControlFlow}
Control flow can be seen as a search method operating within the possible pipeline or feature set spaces. It models the synthesis process and handles the communications between components.

Control flow in pipeline synthesis is implemented through finite-state machines (FSMs). These are built to accept symbolic sequences of node names that act as generic descriptions of all resulting pipelines and effectively represent a search space \emph{specification}. During a FSM transition, the current input symbol is used to query the knowledge system for operation nodes to add, this being done for each of the existing pipelines. The pseudo-code for the FSM transition function is shown in Function~\ref{alg:ps_inner}. Thus, pipelines are built incrementally, independently and in lockstep as linear concatenations of pipeline operations. If the input symbol points to an abstract node, the node hierarchy determines which operation nodes will be considered. The end-result is a set of pipelines that fits both input sequence specifications, node hierarchy as well as the individual node preconditions imposed by the knowledge system. This approach is quite flexible as changes in the FSM input result in new pipeline spaces that are always consistent with the rules encoded in the knowledge system.

In feature synthesis, control flow models the recursive traversal of linked data tables in order to build features, as described in \cite{Kanter2015}. Individual features are assembled in one go, using the template ASTs as blueprints, not incrementally like pipelines. However, the feature set itself is built incrementally as features use recursively previously calculated features as inputs. The pseudo-code from Function~\ref{alg:fs_inner} details how features of a given type are calculated for a specific data table.

\subsection{Program execution (PE)}\label{ssec:ProgramExecution}
Our synthesis approach can be viewed as code generation. For each pipeline or feature set, the code can be compiled and executed. In pipeline synthesis, the execution generates a pipeline object while in feature synthesis, depending on the type of feature, feature functions that follow specific AST templates. If data is available during synthesis, it can be readily passed as input to the pipeline objects or feature functions, providing run-time output information during synthesis.

In both synthesis cases, partial solutions are stored in data structures that hold symbolic node-based representations of ML pipelines and feature ASTs. These are shared and accessible by both control flow and program execution components, allowing therefore the knowledge system to interact in the synthesis dynamically, function of the intermediary data output and symbolic solution structures.

\section{Implementation}\label{sec:Implementation}
From an engineering perspective, we explore an explicitly data-driven implementation. Loosely drawing inspiration from UNIX software design principles \cite{Raymond2003}, the aim is to lift knowledge embedded in programming code to a higher abstraction level. This would allow for a number of emergent advantages such as increased scalability as well as more control over the behavior or functionality of the synthesis processes.

Viewed through this lens, the implementation of the architecture refers to a set of specific data structures and the code that  employs them. At the architectural level this is represented by the existence of an explicit knowledge system. The main advantage here is that by requiring an explicit communication protocol with the knowledge system, a part of the synthesis decisional process is implicitly formalized and becomes transparent. At the component level, the knowledge system has its data component in the form of a knowledge base. Similarly, for the control flow component, in pipeline synthesis the data is the FSM\footnote{This is achievable in practice by separating the requests to the knowledge system and program execution which have a fixed pattern, from the actual FSM transition which can vary function of current state and input.} and in feature synthesis the data are the feature template ASTs.

Arguably, for the program execution module, the data part is represented by pipeline hyper-parameters which can be added in the knowledge base to each of the transforms and learners that support them. This triggers hyper-parameter optimization once a pipeline containing a learner is available. If not present, pipeline components are executed using default parameters either explicitly specified or available in the implementation. To be more exact, the presence of pipeline component hyper-parameters changes the execution of assembled pipeline code rather than the assembly of the code itself.

The rest of the implementation is program code needed to run the FSM, drive feature synthesis and assemble and execute pipelines and features.

The implemented FSM has five states which correspond to stages in the building of a ML pipeline. The inputs correspond to actions that a ML practitioner would take during these stages. Through the FSM, one completely describes the structure of acceptable inputs and by extension, of the pipeline spaces that can be generated as a result.
\begin{figure}[ht]
    \centering
		\begin{tikzpicture}
            [shorten >=2pt,node distance=2cm,on grid,>={Latex[]},auto]

            \node[state,initial]            (nodata)                      {$\mathbf{s}_{\mathrm{\Phi}}$};
            \node[state,accepting]          (data) [right=of nodata] {$\mathbf{s}_{\mathrm{D}}$};
            \node[state,accepting]          (modelabledata) [right=of data] {$\mathbf{s}_{\mathrm{Dm}}$};
            \node[state,accepting]          (model) [right=of modelabledata] {$\mathbf{s}_{\mathrm{M}}$};
            \node[state,accepting]          (end) [right=of model] {$\mathbf{s}_{\mathrm{Me}}$};
            \path[->] (nodata) edge                  node [above]  {$\mathbf{o}_{\mathrm{Ld}}$} (data)
                      (data)   edge  [loop above]    node [above]  {$\mathbf{o}_{\mathrm{Pd}}, \mathbf{o}_{\mathrm{Fd}}, \mathbf{o}_{\mathrm{Dr}}, \mathbf{o}_{\mathrm{DFS}}, \mathbf{o}_{\mathrm{Se}}$} ()
                      (data)   edge                  node [above]  {$\mathbf{o}_{\mathrm{Sm}}$} (modelabledata)
                      (modelabledata)   edge  [loop below]    node [below]  {$\mathbf{o}_{\mathrm{Fd}}, \mathbf{o}_{\mathrm{Dr}}, \mathbf{o}_{\mathrm{DFS}}, \mathbf{o}_{\mathrm{Se}}$} ()
                      (modelabledata)   edge                  node [above]  {$\mathbf{o}_{\mathrm{M}}$} (model)
                      (model)           edge                  node [above]  {$\mathbf{o}_{\mathrm{E}}$} (end);
		\end{tikzpicture}
    \caption{State diagram of the pipeline synthesis FSM. The states and inputs to the FSM have to following meaning: $\mathbf{s}_{\mathrm{\Phi}}$ - no data loaded, $\mathbf{s}_{\mathrm{D}}$ - data loaded, $\mathbf{s}_{\mathrm{Dm}}$ - a model can be applied, $\mathbf{s}_{\mathrm{M}}$- a model has been applied, $\mathbf{s}_{\mathrm{Me}}$ - the model was evaluated; $\mathbf{o}_{\mathrm{Ld}}$ - load data, $\mathbf{o}_{\mathrm{Pd}}$ - pre-process data, $\mathbf{o}_{\mathrm{Fd}}$ - build features, $\mathbf{o}_{\mathrm{Dr}}$ - dimensionality reduction, $\mathbf{o}_{\mathrm{DFS}}$ - deep feature synthesis, $\mathbf{o}_{\mathrm{Sm}}$ - select a model, $\mathbf{o}_{\mathrm{Se}}$ - select a model evaluation method, $\mathbf{o}_{\mathrm{M}}$ - model data, $\mathbf{o}_{\mathrm{E}}$ - evaluate model.}
    \label{img:ps_fsm}
\end{figure}
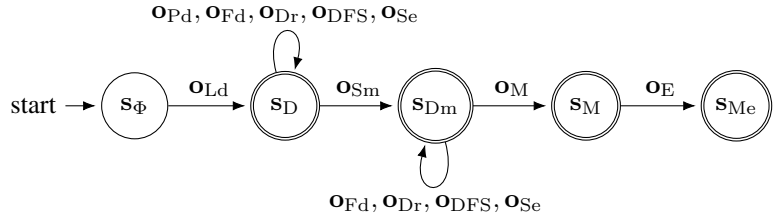
The state diagram shown in Figure~\ref{img:ps_fsm}, illustrates one possible way of modelling the interplay of states and inputs that eventually leads to the creation of pipelines: any succession of data processing operations can be performed between data loading and modelling; in order to apply a model, a model selection operation needs to be specified; the selection of a model evaluation method can occur once data is available however inspection of the pipeline performance can only occur once a model has been trained.

\begin{figure}
    \centering
	\begin{subfigure}{0.25\textwidth}
        \begin{tikzpicture}[level distance=.75cm, sibling distance = 1cm]
            \node {$g_{s}$}
                child { node {$\mathbf{t}$} }
                child { node {$\mathbf{i}$} };
	\end{tikzpicture}
	\caption{'Identity' features}
	\end{subfigure}
	\hfill
    \centering
	\begin{subfigure}{0.25\textwidth}
        \begin{tikzpicture}[level distance=.75cm, sibling distance = 1cm]
            \node {$r_{s}$}
                child { node {$g_{s}$}
                    child { node {$\mathbf{t}$} }
                    child { node {$\mathbf{i}$} }
                }
                child { node {$r_{t}$}
                    child { node {$f_{t}$}
                        child { node {$c_{t}$}}
                            child { node {$g_{t}$}
                                child { node {$\mathbf{t}$} }
                                child { node {$\mathbf{i}$} }
                            }
                        }
                };
	\end{tikzpicture}
	\caption{'Entity' features}
	\end{subfigure}
	\hfill
    \centering
	\begin{subfigure}{0.25\textwidth}
        \begin{tikzpicture}[level distance=.75cm, sibling distance = 1cm]
            \node {$idx$}
                child { node {$g_{t}$}
                    child { node {$\mathbf{t}$} }
                    child { node {$\mathbf{i}$} }
                }
                child { node {$c_{t}$}};
	\end{tikzpicture}
	\caption{'Direct' features}
	\end{subfigure}
	\hfill
    \centering
	\begin{subfigure}{0.25\textwidth}
        \begin{tikzpicture}[level distance=.75cm, sibling distance = 1cm]
            \node {$r_{s}$}
                child { node {$g_{s}$}
                    child { node {$\mathbf{t}$} }
                    child { node {$\mathbf{i}$} }
                }
                child { node {$r_{t}$}
                    child { node {$idx$}
                        child { node {$g_{t}$}
                            child { node {$\mathbf{t}$} }
                            child { node {$\mathbf{i}$} }
                        }
                        child { node {$c_{t}$} }
                    }
                };
	\end{tikzpicture}
	\caption{'Reverse' features}
	\end{subfigure}
    \caption{Feature synthesis AST templates share feature components across feature types. The components correspond to nodes in the knowledge base and have allocated multiple functions: for a single feature type all combinations of feature component functions are instantiated and assembled according to the AST template. The functionality of the components is as follows: $g_s$ and $g_{t}$ represent functions that get and return scalars or tensors respectively and have tensor $\mathbf{t}$ and tensor indices $\mathbf{i}$ as inputs; $r_{s}$ and $r_{t}$ return scalars from one or more input arguments that can be scalars or tensors respectively; $f_{t}$ and $idx$ represent filtering and indexing functions of a tensor using condition functions $c_{t}$ over the tensor.}
    \label{img:fs_asts}
\end{figure}
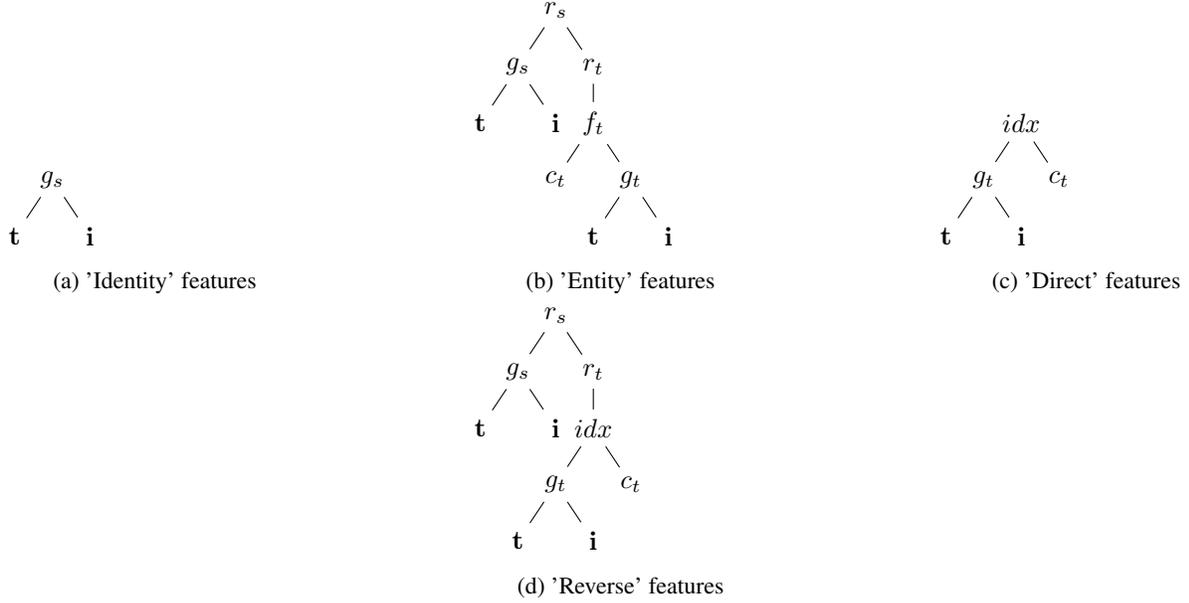

All four feature types described in~\cite{Kanter2015} have been implemented. Their corresponding AST templates are shown in Figure~\ref{img:fs_asts}. Two of them, the 'identity' and 'entity' types will be calculated and employed experimentally as single data tables are used throughout all experiments.

Two knowledge bases have been implemented, for pipeline and feature synthesis respectively, as shown in Figures~\ref{img:kb_pipe_synthesis}~and~\ref{img:kb_feature_synthesis}. The first provides a minimal hierarchy of ML operations and the preconditions associated to them. The knowledge associated to feature synthesis is represented in nodes that have at most two functions allocated, to keep the number of generated features small. The knowledge base source code is stored as editable \texttt{.TOML} files. These are parsed to build property graphs using the \texttt{neo4j}\footnote{Available at \texttt{https://neo4j.com/}} library. The constraint solving functionality is covered by \texttt{CryptoMinSAT}\footnote{Available at \texttt{https://github.com/msoos/cryptominisat}} coupled with a symbolic programming interface\footnote{Available at \texttt{https://github.com/dpsanders/SatisfiabilityInterface.jl}}.

The programming language of choice for the system is \texttt{Julia}~\cite{Bezanson2017}. The system can be seen as a library that uses at its turn an ML library to build pipelines, in this case \texttt{MLJ}\footnote{Available at \texttt{https://github.com/alan-turing-institute/MLJ.jl}~.}~\cite{Blaom2020a, Blaom2020b} and generic Julia libraries to build features. The choice for the ML library was motivated by the fact that \texttt{MLJ} is extensible, integrates many ML algorithms and data transformations, has a good programming model for advanced pipeline composition and supports out of the box hyper-parameter optimization.

\section{Experimental results}\label{sec:ExperimentalResults}
The evaluation of the proposed architecture is designed to validate the main design principles and to shed more light on its functionality. It does not attempt, however, any explicit comparison of the implementation to existing AutoML methods. We considered this necessary at this point for two reasons: first, the scope of the implementation itself is limited strictly to research and hence, not meant to be used as or contend with a production ready library. This translates into a relatively small number of operations implemented in the pipeline and feature synthesis knowledge bases as well as notable computational inefficiencies\footnote{Intermediary synthesis solution data outputs are not cached and partial pipelines have to be completely re-executed when adding new elements if outputs are desired. This is inefficient from a computational standpoint, especially when employing deep feature synthesis in the pipelines. Memory usage is not optimized either to reflect the fact pipelines may share the same operations and data outputs.}. Second, we are confronted with difficulties in choosing the scope of a benchmark comparison. Current AutoML literature focuses on the speed of the synthesis process, accuracy of resulting pipelines and their structure. While this provides a means of ranking AutoML algorithms in an abstract accuracy-computational efficiency space, such an evaluation would not only bring little information regarding the interesting features of our work but also provide little meaningful information of the impact of measured differences in practical real-world applications. These are acknowledged issues in evaluation design in machine learning research \cite{Wagstaff2012} and hold for AutoML as well. Any meaningful solution to this aspect such as grounding the performance in real-world impact metrics is at this point beyond the scope of this work.

\subsection{Pipeline and feature synthesis for the 'XOR' dataset}\label{ssec:XORexperiment}
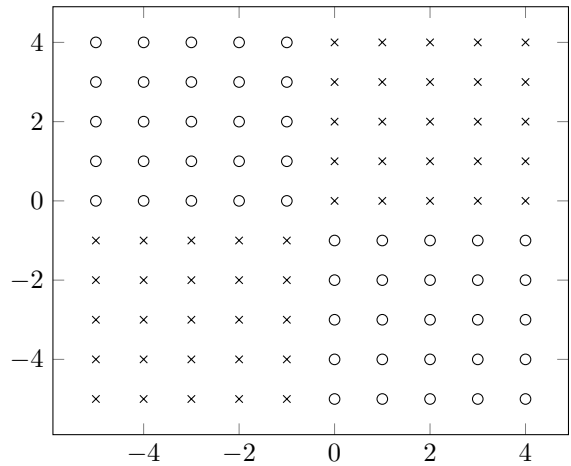
\begin{figure}[ht]
    \centering
    \begin{tikzpicture}
        \begin{axis}[%
                scatter/classes={%
                    1={mark=x,draw=black},
                    0={mark=o,draw=black}}]
            \addplot[scatter,only marks,%
                scatter src=explicit symbolic]%
            table[meta=label] {
            x 	 y 	 label
            -5	-5	1
            -5	-4	1
            -5	-3	1
            -5	-2	1
            -5	-1	1
            -5	0	0
            -5	1	0
            -5	2	0
            -5	3	0
            -5	4	0
            -4	-5	1
            -4	-4	1
            -4	-3	1
            -4	-2	1
            -4	-1	1
            -4	0	0
            -4	1	0
            -4	2	0
            -4	3	0
            -4	4	0
            -3	-5	1
            -3	-4	1
            -3	-3	1
            -3	-2	1
            -3	-1	1
            -3	0	0
            -3	1	0
            -3	2	0
            -3	3	0
            -3	4	0
            -2	-5	1
            -2	-4	1
            -2	-3	1
            -2	-2	1
            -2	-1	1
            -2	0	0
            -2	1	0
            -2	2	0
            -2	3	0
            -2	4	0
            -1	-5	1
            -1	-4	1
            -1	-3	1
            -1	-2	1
            -1	-1	1
            -1	0	0
            -1	1	0
            -1	2	0
            -1	3	0
            -1	4	0
            0	-5	0
            0	-4	0
            0	-3	0
            0	-2	0
            0	-1	0
            0	0	1
            0	1	1
            0	2	1
            0	3	1
            0	4	1
            1	-5	0
            1	-4	0
            1	-3	0
            1	-2	0
            1	-1	0
            1	0	1
            1	1	1
            1	2	1
            1	3	1
            1	4	1
            2	-5	0
            2	-4	0
            2	-3	0
            2	-2	0
            2	-1	0
            2	0	1
            2	1	1
            2	2	1
            2	3	1
            2	4	1
            3	-5	0
            3	-4	0
            3	-3	0
            3	-2	0
            3	-1	0
            3	0	1
            3	1	1
            3	2	1
            3	3	1
            3	4	1
            4	-5	0
            4	-4	0
            4	-3	0
            4	-2	0
            4	-1	0
            4	0	1
            4	1	1
            4	2	1
            4	3	1
            4	4	1
            };
        \end{axis}
    \end{tikzpicture}
    \caption{The 'XOR' dataset used in our experiment. Its two features represent horizontal and vertical integer coordinates each ranged -5 to 4. The data labels \texttt{'$\circ$'} and \texttt{'$\times$'} are arranged so as to match different and identical coordinate sign values respectively. Zero coordinate values are shared between classes.}
    \label{img:xor_dataset}
\end{figure}

This experiment focuses on the interplay of pipeline operations, preconditions and deep feature synthesis. It consists in one execution the synthesis process and the analysis of the process' events and implications for the pipeline space. The problem chosen is the well known 'XOR' toy non-linear classification problem. We use a small dataset of 100 samples with no noise, shown in Figure~\ref{img:xor_dataset}.

We explore domain knowledge integration in two simple yet effective ways: first, for feature synthesis, by allocating a product function to the feature component $r_{t}$ of the 'entity' feature type AST template. This allows the generation of a feature that calculates the product of all existing features across table's lines, enabling perfect separability of the two classes along that dimension. Therefore, simpler linear classifiers could be employed whenever such a feature is present. Therefore, as a second way of integrating domain knowledge, a linear separability precondition is added in the pipeline synthesis knowledge base. It enables the use of linear classifiers whenever data is linearly separable. Moreover, a second precondition that enables only non-linear classifiers if the reverse is true is added as well.

\begin{figure}
    \centering
    \includegraphics[width=\textwidth]{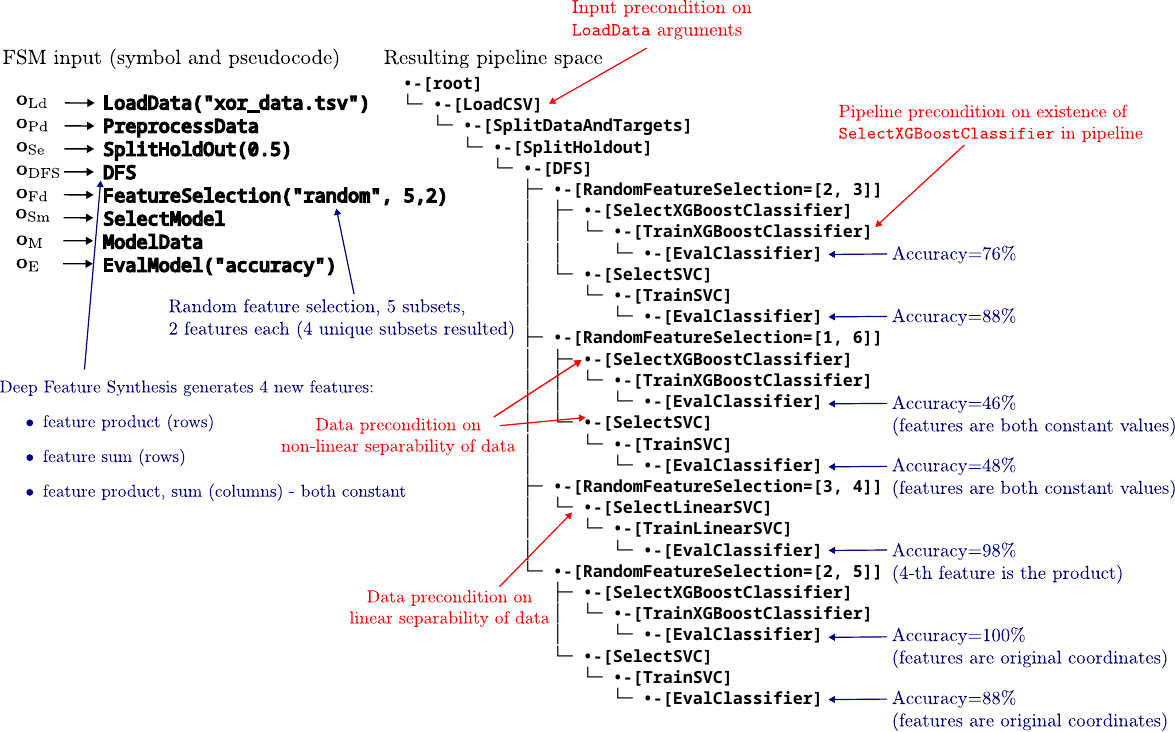}
    \caption{'XOR' problem FSM input and resulting pipeline space. Individual pipeline accuracy is highest if the features are informative and does not depend on the classifier used. Preconditions are being used to enable specific pipeline operations by processing information from the FSM input, pipeline structures or by testing the pipeline outputs.}\label{img:xor_experiment_results}
\end{figure}

Figure~\ref{img:xor_experiment_results} shows the input to the synthesis algorithms, the resulting pipeline space and highlights other information relevant to the synthesis process that can be monitored or extracted from the logs or synthesis outputs. Deep feature synthesis is employed as a standalone operation and added to the pipelines. The accuracies reported have been calculated on a test set randomly sampled to amount for 50\% of the data.
New pipelines are generated when more than one classifier is available or when data is split randomly into feature subsets. During synthesis, not all pipeline nodes contribute with actual code to the pipeline generating program. An example is model selection, which serves to indicate which specific model has been selected and informs preconditions linked to training nodes. The effect of domain knowledge integration is noticeable in that the linear classifier is selected (\texttt{SelectLinearSVC}) and trained (\texttt{TrainLinearSVC}) only if the feature with index \texttt{4} in the table is used, which corresponds to the product feature. The lowest accuracies are around 50\% and due to the fact that some deep features are constant\footnote{The classifiers assign all samples from the balanced test dataset to a single class.}.

\subsection{Multiple pipeline syntheses for the 'Circles' dataset}\label{ssec:Circlesexperiment}
\begin{figure}[ht]
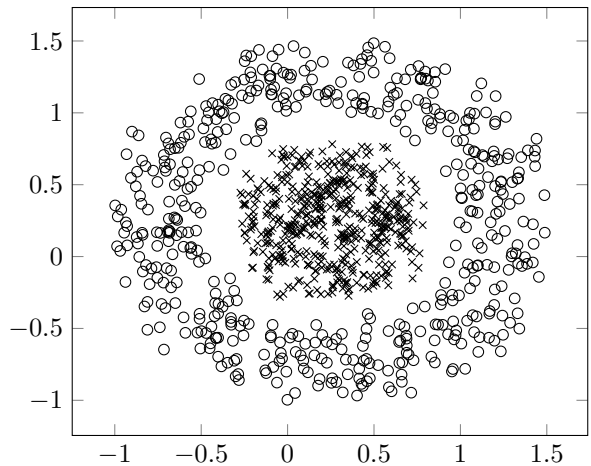

    \centering

    \caption{The 'circles' dataset used in our experiment. Its two features correspond to 'noisy' floating point horizontal and vertical coordinates of two concentric circular point clouds of different radii, marked by the labels \texttt{'$\circ$'} and \texttt{'$\times$'} respectively.}
    \label{img:circles_dataset}
\end{figure}

This experiment focuses on multiple runs of the synthesis process, using different inputs and preconditions. Random feature selection is not used and as a result, synthesis outputs depend only on FSM inputs and the knowledge base. The dataset, shown in Figure~\ref{img:circles_dataset}, has 1000 samples and contains two concentric circular point clouds.

The output pipeline space sizes, error properties across their pipelines as well as the corresponding inputs are shown in Table~\ref{table:circles_experiment}. There are nine operations implemented in the knowledge base for classifying, generating features and performing dimensionality reduction. These are shown in Table~\ref{app:table:ops_and_preconditions} along with other operations and preconditions that can reduce the pipeline space.
\begin{table}[ht]
\begin{center}
    \begin{minipage}{\textwidth}
    \caption{The FSM input structure and data precondition impact on pipeline space characteristics.}\label{table:circles_experiment}%
    \centering
    \begin{tabular}{@{}lccccc@{}}
        \toprule
            FSM input\footnotemark[1] & active data preconditions & \# & $\textrm{\#}_{\textrm{max}}$\footnotemark[2] & mean $\mathrm{logloss}$ & stddev. $\mathrm{logloss}$\\
        \midrule
            $\mathbf{o}_{\mathrm{Sm}}$                                                                                         & all  & 2   & 3   & 0.011  & 0.016  \\
            $\mathbf{o}_{\mathrm{Fd}}$ , $\mathbf{o}_{\mathrm{Sm}}$                                                            & all  & 6   & 9   & 0.007  & 0.009  \\
            $\mathbf{o}_{\mathrm{Dr}}$ , $\mathbf{o}_{\mathrm{Sm}}$                                                            & all  & 3   & 9   & 0.149  & 0.246  \\
            $\mathbf{o}_{\mathrm{Fd}}$ , $\mathbf{o}_{\mathrm{Dr}}$ , $\mathbf{o}_{\mathrm{Sm}}$                               & all  & 12  & 27  & 0.084  & 0.205  \\
            $\mathbf{o}_{\mathrm{Fd}}$ , $\mathbf{o}_{\mathrm{Fd}}$ , $\mathbf{o}_{\mathrm{Dr}}$ , $\mathbf{o}_{\mathrm{Sm}}$  & all  & 34  & 81  & 0.075  & 0.154  \\
            & & & & & \\
            $\mathbf{o}_{\mathrm{Sm}}$                                                                                         & none & 3   & 3   & 5.964  & 10.318 \\
            $\mathbf{o}_{\mathrm{Fd}}$ , $\mathbf{o}_{\mathrm{Sm}}$                                                            & none & 9   & 9   & 6.767  & 10.356 \\
            $\mathbf{o}_{\mathrm{Dr}}$ , $\mathbf{o}_{\mathrm{Sm}}$                                                            & none & 9   & 9   & 3.602  & 7.022  \\
            $\mathbf{o}_{\mathrm{Fd}}$ , $\mathbf{o}_{\mathrm{Dr}}$ , $\mathbf{o}_{\mathrm{Sm}}$                               & none & 24  & 27  & 4.218  & 7.471  \\
            $\mathbf{o}_{\mathrm{Fd}}$ , $\mathbf{o}_{\mathrm{Fd}}$ , $\mathbf{o}_{\mathrm{Dr}}$ , $\mathbf{o}_{\mathrm{Sm}}$  & none & 60  & 81  & 4.850  & 8.271  \\
    \end{tabular}
    \footnotetext[1]{Only the FSM input symbols which correspond to abstract operations that increase the pipeline space are shown. In practice, each sequence is preceded by $\mathbf{o}_{\mathrm{Ld}}$, $\mathbf{o}_{\mathrm{Pd}}$ and $\mathbf{o}_{\mathrm{Se}}$ and followed by $\mathbf{o}_{\mathrm{M}}$ and $\mathbf{o}_{\mathrm{E}}$}
    \footnotetext[2]{Theoretical maximum number of pipelines, given the number of operation nodes available.}
    \end{minipage}
\end{center}
\end{table}

From the results it becomes apparent that disabling the preconditions that operate on the partial pipeline outputs increases the both errors and number of pipelines. This increased error is due to the fact that the linear classifier is used even when data is not linearly separable which leads to larger errors across all pipeline spaces. The larger number of pipelines is due to the fact that in this implementation data-focused preconditions have a larger impact in the reduction of the pipeline space than the pipeline structure-focused ones.

\section{Discussion}\label{sec:Discussion}


Our work differs from existing research in that we explored an architecture with the goal to improve domain knowledge integration and explain-ability. Both can be viewed as emergent properties of the explicit knowledge-driven architectural approach rather than features hard-coded in the implementation. Yet, some of the aspects of the system depend largely on the quality and size of the knowledge itself. This can be observed in the second experiment where preconditions that operate on pipeline outputs have a larger impact on the final pipeline space that other types of preconditions. This is by no means a guarantee for other datasets as the same preconditions may not apply. This hints at the previously mentioned difficulty in comparing ML algorithms and points to the importance of contextualizing the evaluation when employing domain knowledge - if one is to bring any interpretation of results to bear, it must in some way be connected to or reflect the application or context in which the system operates.

It is interesting to note that the work also addresses, albeit indirectly, other important issues that AutoML currently faces~\cite{Zoller2021}, namely, relatively unsophisticated solutions with respect to human ones, a narrow focus of designs on specific ML problems such as supervised learning and little integration with feature engineering approaches.

We recognize that AutoML is difficult to approach in a holistic way and design trade-offs are unavoidable. This approach is no exception and comes with some difficulties and challenges of its own. Our vision followed the principles of simplicity in implementation, modularity in design and the assumption of a mixed, networked context. These principles would translate into systems made of simple, interconnected components that employ protocols to communicate with each other and that exist within a greater analytical landscape, where inputs, outputs and connected data structures are accessible to humans and machine users.

The presence of the knowledge base itself requires some actual representation and integration of specific ML and data science domain knowledge. Interfacing with the knowledge system and observing its communication allows to explain the decisions that the system takes in building pipelines and features. Even if both experiments presented here are classification tasks, the system is by no means specialized to it and can produce relatively complex pipelines for other related tasks, such as unsupervised learning or visualization, in a straightforward manner.

We noticed that if an intuitive and easy way to access and update the knowledge-base is available, scalability and customization by manual means become naturally approachable as new operations, preconditions and feature components can be added without modifications to the synthesis program code.

Hyper-parameter optimization is handled out of the box, as a consequence of using \texttt{MLJ} as a ML library back-end, and decoupled from the pipeline component selection mechanism. Although it is difficult to state what is the impact of the accuracy or complexity of the final pipelines, the system benefits in terms of simplicity and transparency of the mechanism: hyper-parameter-related information for ML operations is present in knowledge base and enables optimization, once the pipelines contain learners.

Relying on a knowledge source may incur additional efforts in the initial development and understanding of the implementation. More importantly, conceptual errors in the knowledge design phase can have negative consequences during synthesis, resulting in logically erroneous yet operational pipelines.

The sources of software bugs can become difficult to find as there are several layers of abstraction and code execution present. This was an important issue encountered during development: bugs in the code of knowledge base nodes were notoriously subtle and hard to track.

Estimating the complexity or, results space size, of the synthesis process remains challenging as well, as it depends on the data characteristics and types of preconditions present. Generally, experimentation in this direction using smaller datasets and avoiding slow operations such as hyper-parameter optimization can provide some indication however significant practitioners' experience is necessary. We did not attempt to implement mechanisms to constrain the syntheses as to perform within a given computational resource budget and this remains an open issue.

Practical evaluation in a real application should be however more straightforward, even though subjective by nature. This is because results depend on the knowledge base which at its turn is assumed to have been developed to reflect one or more ML practitioners' insights, algorithms and domain knowledge.

At this point it is unclear to what extent a knowledge system can be decoupled from a specific ML library and programming language. This may imply ML library development adhering or supporting a given AutoML interface requirement, which is not the case currently. One possible limitation is that the functioning of the knowledge system and the way it interacts with the rest of the system must be formalized and documented in order adequately use and maintain it. The implemented system relies on implicit semantics in the representation system which proved itself quite practical given the scope of our work. A production-ready implementation may adopt a formal representation system however, this poses not only a compromise between expressivity and tractability\cite{Levesque1984, Brachman1984, Patterson2017-2} but most probably would require also significant implementation and maintenance efforts. Recent progress in this aspect~\cite{Patterson2017, Patterson2017-2, Patterson2018} provides insight into what could be possible in this direction.

The architecture was envisioned to communicate with other ML or analytical systems, scale in solution complexity and reduce programming efforts. Because one can provide as input arbitrarily abstract pipeline operations, the approach is well suited for more targeted exploration of the pipeline space, in which, for example, several key algorithms or operations are fixed. The FSM input could, in such cases, be provided by either automated systems that parse or mine data science experiments or, through human-in-the-loop approaches~\cite{Russell2019}. One could also exploit automatic or human-driven knowledge base completion mechanisms. This would imply learning specific code transformations that would allow code snippets from third-party code repositories to be automatically adapted for use during synthesis, approach that is to some extent used in NAS contexts~\cite{Peng2020}.

Scalability in solution sophistication can be approached in several ways, such as creating pipeline ensembles or, by building pipelines considering code assembly paradigms beyond linear concatenation~\cite{Nikitin2021}. This would require effective use of additional knowledge regarding the ways and circumstances in which pipelines and nodes can be assembled in order to effectively move beyond the linear concatenation of nodes.

Lastly, multi-objective optimization in the AutoML context~\cite{Elsken2018, Pfisterer2019}, could also be pursued explicitly by employing preconditions that analyze the assembled code complexity or pipeline runtime inference speed. This is to some extent the case already as classifier selection based on data structure i.e. using a linear classifier when data is linearly separable, favors simplicity, which constitutes a complexity optimization of the pipeline.

\section{Conclusion}\label{sec:Conclusion}
In this paper we propose and explore a knowledge-driven architecture for AutoML which aims to improve domain knowledge integration and explain-ability. The architecture is used to implement a reference, minimalistic, data-driven system for ML pipeline and data feature generation. The system is tested in two experimental scenarios, designed to provide an early view on the potential of the approach. We also discuss the overall vision behind the approach, trade-offs and future potential that the architecture has to offer. In conclusion, we strongly believe that the ideas presented here are a necessary step towards the development of more explainable and scalable AutoML solutions and rendering them integral parts of larger ML software ecosystems.

\section*{Acknowledgements}\label{sec:Acknolwedgements}
This research was financially supported by  Vlaanderen Agentschap Innoveren \& Ondernemen (VLAIO), Project~{HBC.2018.2016} in collaboration with Omina Technologies bvba.
\clearpage

\appendix
\section{Appendices}

\subsection{Operations and preconditions}\label{app:sec:Operations_preconditions}
\begin{table}[ht]
\begin{center}
    \begin{minipage}{\textwidth}
    \centering
    \caption{Description of operation nodes and linked preconditions for the 'circles' dataset experiment.}\label{app:table:ops_and_preconditions}%
    \begin{tabular}{@{}lll@{}}
        \toprule
            FSM symbol & operation nodes & linked precondition(s)\footnotemark[1]\\
            \midrule
            \multirow{3}*{$\mathbf{o}_{\mathrm{Sm}}$}
                        & selects linear SVM & Data is linearly separable\footnotemark[2] \\
                        \cline{2-3}
                        & selects non-linear SVM & Data is not linearly separable \\
                        \cline{2-3}
                        & selects XGBoost & Data is not linearly separable \\
            \midrule
            \multirow{3}*{$\mathbf{o}_{\mathrm{M}}$}
                        & trains linear SVM & linear SVM selected \\
                        \cline{2-3}
                        & trains  non-linear SVM& non-linear SVM selected \\
                        \cline{2-3}
                        & trains XGBoost & XGBoost selected \\
            \midrule
            \multirow{3}*{$\mathbf{o}_{\mathrm{Fd}}$}
                        & NOP (no operation)& - \\
                        \cline{2-3}
                        & Feature kernel & - \\
                        \cline{2-3}
                        & Feature product & Previous operation is not a feature kernel \\
            \midrule
            \multirow{3}*{$\mathbf{o}_{\mathrm{Dr}}$}
                        & NOP (no operation) & - \\
                        \cline{2-3}
                        & PCA\footnotemark[3] & Input data has more than 2 features \\
                        \cline{2-3}
                        & Kernel PCA\footnotemark[4] & A feature kernel is not in the current pipeline \\
    \end{tabular}
    \footnotetext[1]{Only the preconditions that influence the pipeline space are listed.}
    \footnotetext[2]{This is done by applying LDA (linear discriminant analysis) on the dataset and checking its classification performance against a preconfigured threshold.}
    \footnotetext[3]{Output dimensionality is fixed at 3 components.}
    \footnotetext[4]{Output dimensionality is fixed at 2 components. The RBF kernel tweaked so that the transformed dataset is linearly separable.}
    \end{minipage}
\end{center}
\end{table}

\clearpage

\subsection{Algorithms}\label{app:sec:Additional_Algorithms}


\begin{algorithm}
    \floatname{algorithm}{Function}
    \caption{Knowledge system query answering}
    \label{alg:ks_query}
    \begin{algorithmic}[1]
        \Require {Knowledge system $KS$ and synthesis data $synthesis\_data$}
        \Ensure {Operation nodes $reply$}
        \Function{ks\_query}{$KS, synthesis\_data$}
            \State $KB \leftarrow <\textrm{knowledge base of KS}>$ \Comment{Database connection}
            \State $CS \leftarrow <\textrm{constraint solver of KS}>$
            \State $query \leftarrow \textsc{transform\_to\_neo4j}(synthesis\_data)$
            \State $data\_kb \leftarrow \textsc{kb\_query}(KB, query)$ \Comment{Get node and precondition symbols, code}
            \State $solution \leftarrow \textsc{build\_and\_solve\_csp}(CS, synthesis\_data, data\_kb)$ \Comment{Synthesis data is necessary for precondition execution}
            \State $reply \leftarrow \textsc{add\_data}(solution, data\_kb)$ \Comment{Re-maps to KB symbols and code}
            \State $\textbf{return } reply$
        \EndFunction
    \end{algorithmic}
\end{algorithm}

\begin{algorithm}
    \floatname{algorithm}{Function}
    \caption{Constraint satisfaction program building and solving}
    \label{alg:cs_build_solve}
    \begin{algorithmic}[1]
        \Require {Knowledge system $KS$, synthesis data $synthesis\_data$, node and precondition symbols and code $data\_kb$}
        \Ensure {Constrain solver output $solution$}
        \Function{build\_and\_solve\_csp}{$KS, synthesis\_data, data\_kb$}
            \State $constraints \leftarrow \emptyset$
            \For{$node\_list \in data\_kb$} \Comment{A node list corresponds to all possible options for one pipeline operation or feature component}
                \For{$n \in node\_list$}
                    \State $s \leftarrow <\textrm{a random symbol (mapping is stored)}>$
                    \State $constraints \leftarrow constraints \cup (s \in \{0,1\})$
                    \For{$p \in \textsc{preconditions}(data\_kb, n)$}
                        \State $r \leftarrow \textsc{execute}(p, synthesis\_data)$
                        \State $constraints \leftarrow constraints \cup (s = s \wedge r)$
                    \EndFor
                \EndFor
                \State $constraints \leftarrow constraints \cup (\sum_{nodes\_list}\left(n\right) = 1)$
            \EndFor
            \State $solution \leftarrow \textsc{solve}(KS, constraints)$ \Comment{Call to CSP solver (CryptoMinSat)}
            \State $\textbf{return } solution$
        \EndFunction
    \end{algorithmic}
\end{algorithm}

\begin{algorithm}
    \floatname{algorithm}{Function}
    \caption{Pipeline synthesis {FSM} transition function}
    \label{alg:ps_inner}
    \begin{algorithmic}[1]
        \Require {FSM input state $S$, input symbol $O$, knowledge system $KS$ and partial solution $Pipelines$}
        \Ensure {New FSM state $S'$}
        \Function{Transition}{$S, O, KS, Pipelines$}
            \For{$pipe \in Pipelines$}
                \State $synthesis\_data \leftarrow \left(O, pipe,\;<\textrm{data outputs of}\;pipe>\right)$
                \State $node\_list \leftarrow \textsc{ks\_query}(KS, synthesis\_data)$
                \For{$n \in node\_list$}
                    \State $new\_pipe \leftarrow pipe\; \cup\; n$
                    \If{$O \textrm{ is executable}$}
                        \State $\textsc{store}(new\_pipe, \textsc{execute}(new\_pipe))$
                    \EndIf
                    \State $Pipelines \leftarrow Pipelines\; \cup\; new\_pipe$ \Comment{Solution update}
                \EndFor
                \State $\textsc{delete}(pipe)$  \Comment{'Old' pipeline no longer necessary}
            \EndFor
            \State $S' \leftarrow \textsc{fsm\_transition}(S,O)$ \Comment{Next {FSM} state based on }
            \State $\textbf{return } S'$
        \EndFunction
    \end{algorithmic}
\end{algorithm}

\begin{algorithm}
    \floatname{algorithm}{Function}
    \caption{Feature synthesis function}
    \label{alg:fs_inner}
    \begin{algorithmic}[1]
        \Require {Type of features to calculate $feature\_type$, all linked tables and current table $table\_data$, knowledge system $KS$ and partial solution $Features$}
        \Ensure {Updated partial solution $Features$}
        \Function{Build\_Features}{$feature\_type, table\_data, KS, Features$} \Comment{data contains all data tables, links and current table pointer}
            \State $table \leftarrow <\textrm{data table from which data is read}>$
            \For{$column \in table$}
                \State $synthesis\_data \leftarrow \left(feature\_type, table\_data, column\right)$ \Comment{Synthesis data}
                \State $components\_list \leftarrow \textsc{ks\_query}(KS, synthesis\_data)$
                \State $t\_ast \leftarrow \textsc{ast\_template}(feature\_type)$ \Comment{A feature AST template}
                \For{$components \in components\_lists$}
                    \State $name \leftarrow <\textrm{a random symbol}>$
                    \State $f\_ast \leftarrow \textsc{materialize}(t\_ast, components)$ \Comment{Executable code {AST}}
                    \If{features are to be calculated}
                        \State $vals \leftarrow \textsc{execute}(f\_ast, synthesis\_data)$  \Comment{Feature input is synthesis data}
                    \Else \State $vals \leftarrow \emptyset$
                    \EndIf
                    \State $feature \leftarrow \textsc{build}(f\_ast, vals, name)$ \Comment{Features are objects}
                    \State $Features \leftarrow Features \cup feature$
                \EndFor
            \EndFor
            \State $\textbf{return } Features$
        \EndFunction
    \end{algorithmic}
\end{algorithm}

\clearpage

\subsection{Knowledge structures}

\begin{figure}[ht]
    \centering
    \includegraphics[width=0.6\textwidth]{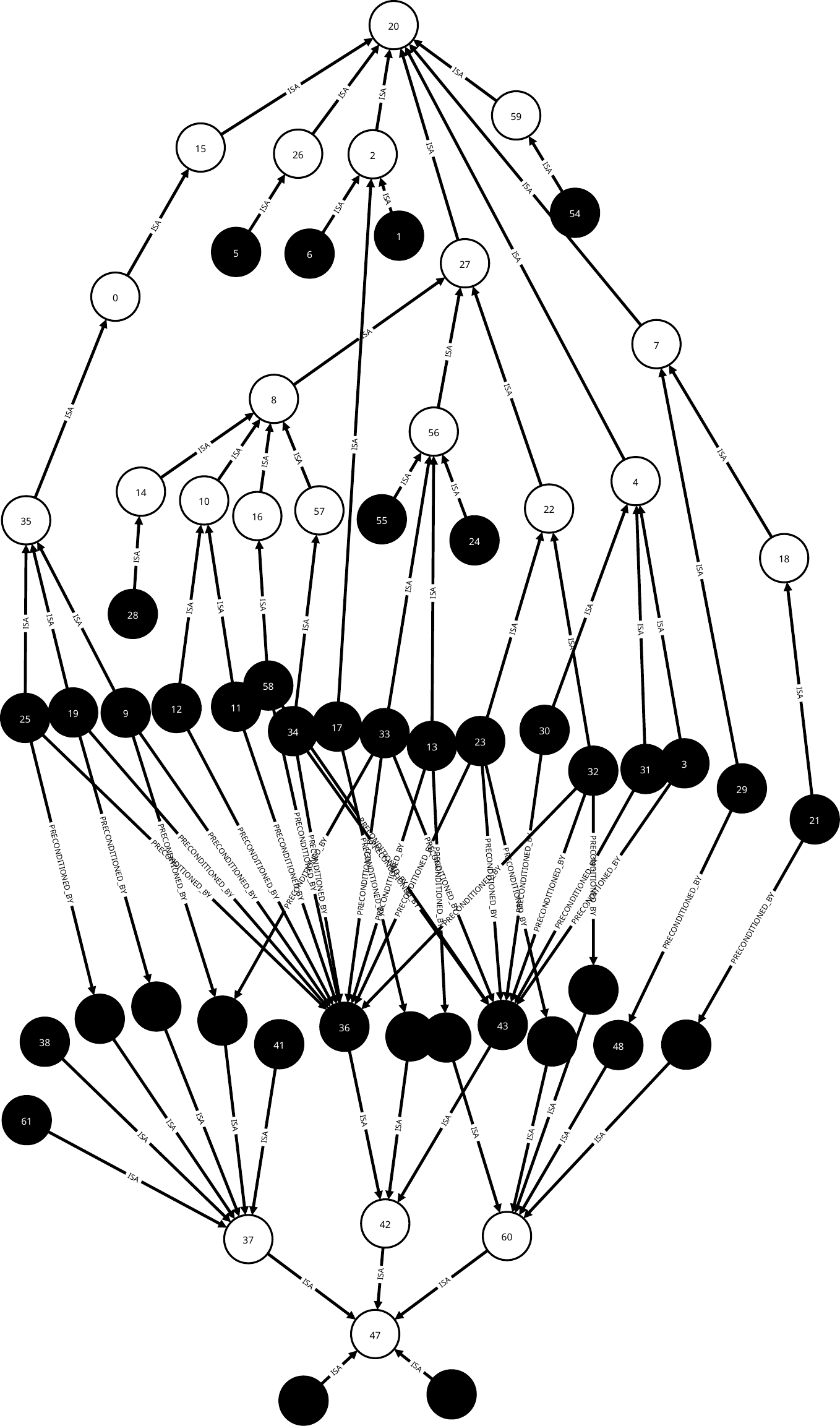}
    \caption{The pipeline synthesis knowledge organization implemented in \texttt{neo4j}. From top to bottom, one can observe a node hierarchy made of abstract nodes (white), operation nodes (black) and hierarchical links (\texttt{ISA}), precondition links (\texttt{PRECONDITIONED\_BY}) and finally, preconditions (black). In the implementation, the preconditions are also organized in a hierarchy - identifiable by the bottom white vertices - which is used to enable or disable certain types of preconditions.}\label{img:kb_pipe_synthesis}
\end{figure}

\begin{figure}[ht]
    \centering
    \includegraphics[width=0.6\textwidth]{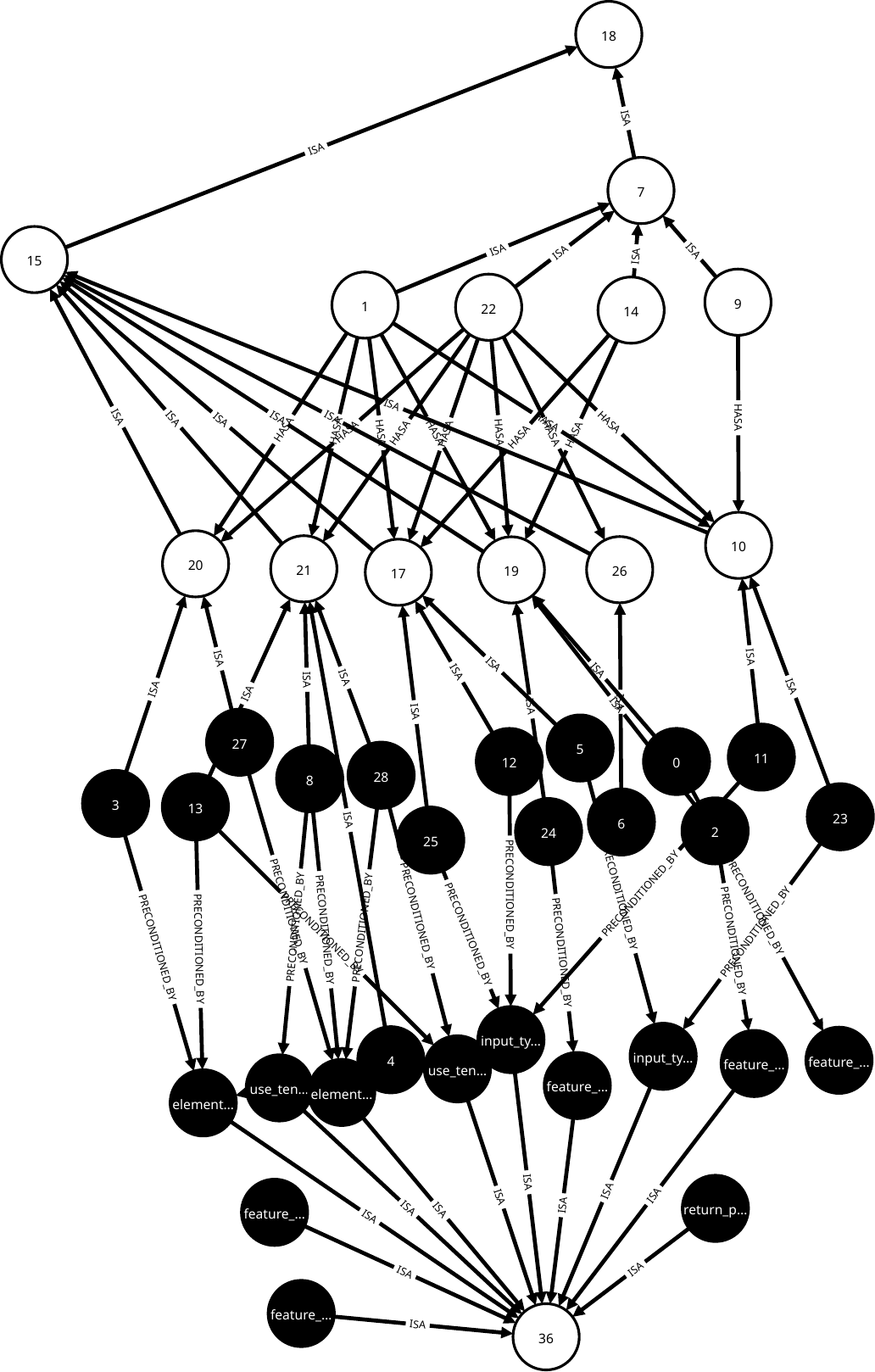}
    \caption{The feature synthesis knowledge organization implemented in \texttt{neo4j}. From top to bottom, one can observe a node hierarchy made of abstract nodes (white), operation nodes (black), hierarchical links (\texttt{ISA}) and feature AST links (\texttt{HASA}), precondition links (\texttt{PRECONDITIONED\_BY}) and finally, preconditions (black).}
    \label{img:kb_pipe_synthesis}
\end{figure}

\clearpage

\bibliographystyle{apalike}
\bibliography{kbdaml}

\end{document}